\documentclass[10pt,twocolumn,letterpaper]{article}

\usepackage{cvpr}
\usepackage{times}
\usepackage{epsfig}
\usepackage{graphicx}
\usepackage{epstopdf}
\usepackage{amsmath}
\usepackage{amssymb}
\usepackage{booktabs}  
\usepackage{enumitem}
\usepackage{subfig}
\usepackage{tabularx}
\usepackage{rotating}

\usepackage{booktabs}       
\usepackage{algorithm}
\usepackage{algpseudocode}
\usepackage[square,sort,comma,numbers]{natbib}
\usepackage{color}
\usepackage{pifont} 
\usepackage{physics}
\usepackage{caption} 

\usepackage{nicefrac}      
\usepackage{import}
\usepackage{color}
\usepackage{transparent}
\graphicspath{ {figures/} }

\usepackage[pagebackref=true,breaklinks=true,letterpaper=true,colorlinks,bookmarks=false]{hyperref}

\newcommand{\xmark}{\ding{55}}%
\newcommand\blfootnote[1]{%
  \begingroup
  \renewcommand\thefootnote{}\footnote{#1}%
  \addtocounter{footnote}{-1}%
  \endgroup
}
\usepackage[hang,flushmargin]{footmisc}

\cvprfinalcopy 

\ifcvprfinal\fi
\begin{document}

\title{First-Person Hand Action Benchmark with RGB-D Videos \\and 3D Hand Pose Annotations}
\author{
Guillermo Garcia-Hernando\hspace*{8mm}Shanxin Yuan\hspace*{8mm}Seungryul Baek\hspace*{8mm} Tae-Kyun Kim\\
Imperial College London\\
\small\texttt{\{g.garcia-hernando,s.yuan14,s.baek15,tk.kim\}@imperial.ac.uk}
}

\maketitle

\begin{abstract}
\vspace{-2mm}

In this work we study the use of 3D hand poses to recognize first-person dynamic hand actions interacting with 3D objects. Towards this goal, we collected RGB-D video sequences comprised of more than 100K frames of 45 daily hand action categories, involving 26 different objects in several hand  configurations. To obtain hand pose annotations, we used our own mo-cap system that automatically infers the 3D location of each of the 21 joints of a hand model via 6 magnetic sensors and inverse kinematics. Additionally, we recorded the 6D object poses and provide 3D object models for a subset of hand-object interaction sequences. To the best of our knowledge, this is the first benchmark that enables the study of first-person hand actions with the use of 3D hand poses. We present an extensive experimental evaluation of RGB-D and pose-based action recognition by 18 baselines/state-of-the-art approaches. The impact of using appearance features, poses, and their combinations are measured, and the different training/testing protocols are evaluated. Finally, we assess how ready the 3D hand pose estimation field is when hands are severely occluded by objects in egocentric views and its influence on action recognition. From the results, we see clear benefits of using hand pose as a cue for action recognition compared to other data modalities. Our dataset and experiments can be of interest to communities of 3D hand pose estimation, 6D object pose, and robotics as well as action recognition. 
\end{abstract}
\vspace{-5mm}
\thispagestyle{empty} 

\begin{figure}[!t]
\begin{center}
\def\svgwidth{\columnwidth}
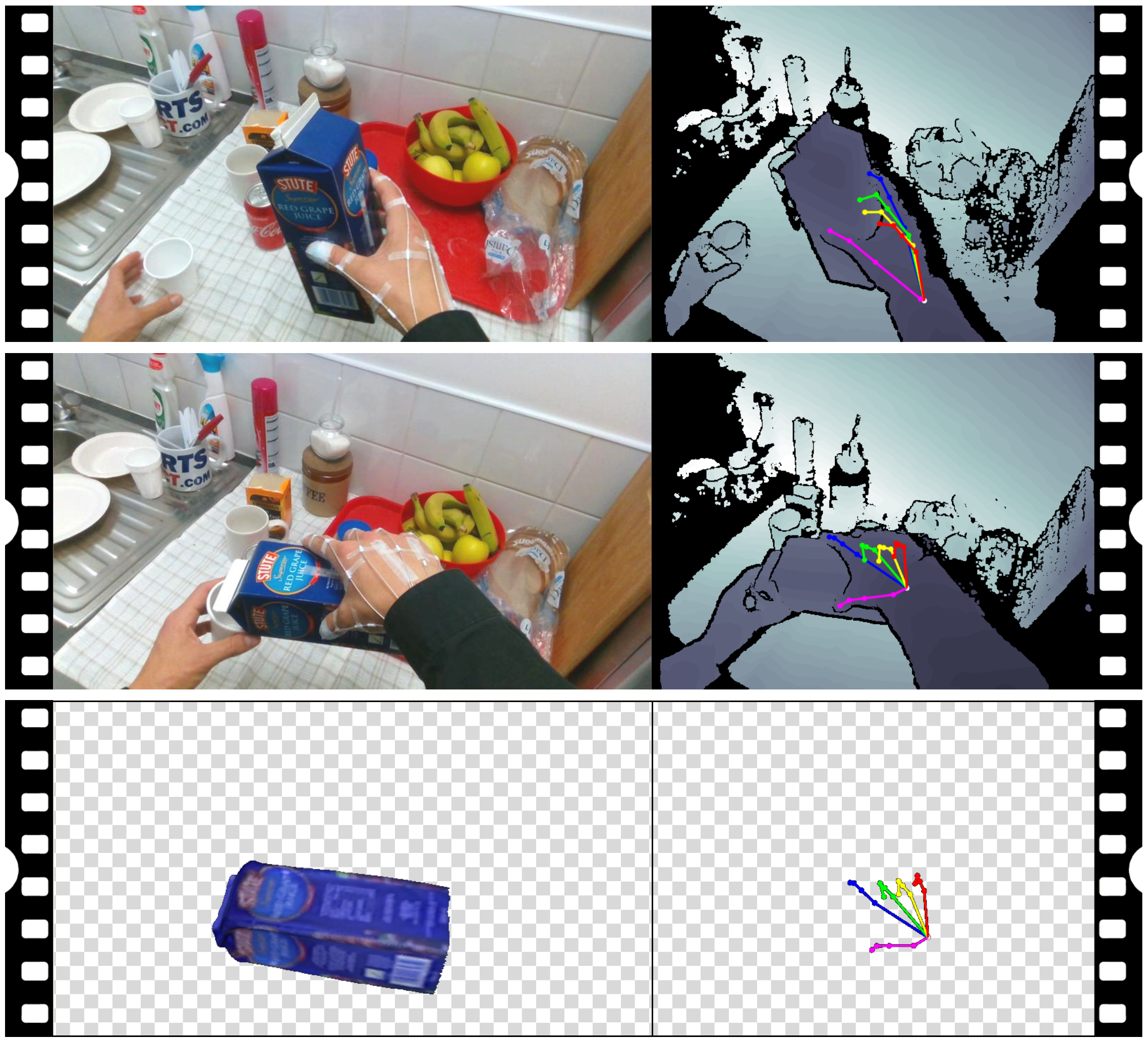
   \caption{We show two frames from a sequence belonging to the action class `pour juice'. We propose a novel first-person action recognition dataset with RGB-D videos and 3D hand pose annotations. We use magnetic sensors and inverse kinematics to capture the hand pose. On the right we see the captured depth image and hand pose. We also captured 6D object pose for a subset of hand-object actions.}
\label{fig:teaser}	
\end{center}
\end{figure}

\section{Introduction}
\blfootnote{Dataset visualization: \url{https://youtu.be/U5gleNWjz44}}
We interact with the world using our hands to manipulate objects, machines, tools, and socialize with other humans. In this work we are interested in understanding how we use our hands while performing daily life \textit{dynamic} actions with the help of fine-grained hand pose features, a problem of interest for multiple applications requiring high precision, such as hand rehabilitation~\cite{allin2007assessment}, virtual/augmented reality \citep{jang20153d}, teleoperation~\cite{zollo2007biomechatronic}, and robot imitation learning \cite{argall2009survey}. 

Previous work in first-person action recognition \cite{ishihara2015recognizing, caiunderstanding,ma2016going,singh2016first} found that daily actions are well explained by looking at hands, a similar observation found in third-person view \cite{yang2015grasp}. In these approaches, hand information is extracted from hand silhouettes \cite{ma2016going,singh2016first} or discrete grasp classification \cite{ishihara2015recognizing, caiunderstanding, rogez2015understanding} using low-level image features. 
In full-body human action recognition it is known that using higher level and viewpoint invariant features such as body pose can benefit action recognition \cite{yao2011does,wu2014leveraging,zhangefficient,yu2013unconstrained, shi2017learning}, although this has not yet been studied in detail for hands. Compared to full-body actions, hand actions present unique differences that make the use of pose as a cue not obvious: style and speed variations across subjects are more pronounced due to a higher degree of mobility of fingers and the motion can be very subtle. A setback for using hand pose for action recognition is the absence of reliable pose estimators off-the-shelf in contrast to full body \cite{shotton2013real,wei2016cpm}, mainly due to the absence of hand pose annotations on real (\textit{cf.} synthetic) data sequences, notably when objects are involved~ \cite{rogez20143d,rogez2015first,mueller2017real,choi2017robust}.

In this work we introduce a new dataset of first-person dynamic hand action sequences with more than 100,000 RGB-D frames annotated with 3D hand poses, using six magnetic sensors attached to the fingertips and inverse kinematics. We captured 1175 action samples including 45 categories manipulating 26 different objects in 3 scenarios. We designed our hand actions and selected objects to cover multiple hand configurations and temporal dynamics. Furthermore, to encourage further research, we also provide 6-dimensional object pose ground truth, and their 3D mesh models, for 4 objects spanning, 10 different actions. We evaluate several baselines and state-of-the-art RGB-D and pose-based action recognition in our dataset and test the current state-of-the-art in hand pose estimation and its influence on action recognition. To the best of our knowledge, this is the first work that studies the problem of first-person action recognition with the use of hand pose features and the first benchmark of its kind. In summary, the contribution of this paper is three-fold: 

\textbf{Dataset:} we propose a fully annotated dataset to help the study of egocentric dynamic hand-object actions and poses. This is the first dataset to combine both fields in the context of hands in real videos and quality hand pose labels. 

\textbf{Action recognition:} we evaluate 18 baselines and state-of-the-art approaches in RGB-D and pose-based action recognition using our proposed dataset. Our selected methods cover most of the research trends in both methodology and use of different data modalities.

\textbf{Hand pose:} We evaluate a state-of-the-art hand pose estimator in our real dataset, \textit{i.e.}, the occluded setting of hand-object manipulations and assess its performance for action recognition.

\section{Related work}

\begin{figure*}[!t]
\includegraphics[width=\textwidth]{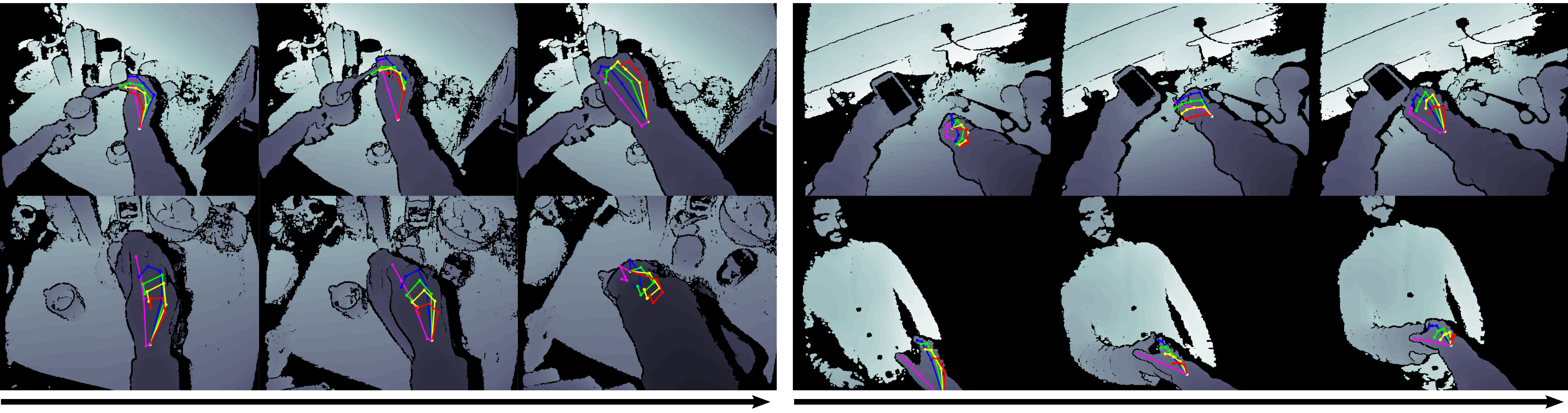}
   \caption{\textbf{Hand actions:} We captured daily hand actions using a RGB-D sensor and used a mo-cap system to annotate hand pose. \textbf{Left:}  `put sugar' and `pour milk' (kitchen). \textbf{Right:} `charge cell phone' (office) and `handshake' (social). }
\label{fig:dataset}	
\end{figure*}

\textbf{Egocentric vision and manipulations datasets:} The important role of hands while manipulating objects has attracted the interest from both computer vision and robotics communities. From an action recognition perspective and only using RGB cues, recent research \cite{fathi2011understanding,gtea,pirsiavash2012detecting,ma2016going,singh2016first,bambach2015lending} has delved into recognizing daily actions and determined that both manipulated objects and hands are important cues to the action recognition problem. A related line of work is the study of human grasp from a robotics perspective~\cite{bullock2015yale,cai2015scalable}, as a cue for action recognition \cite{yang2015grasp, ishihara2015recognizing, caiunderstanding,fermuller2017prediction}, force estimation \cite{kry2006interaction,rogez2015understanding,fermuller2017prediction}, and as a recognition problem itself~\cite{huang2015we,rogez2015understanding}. Recently, \cite{luoscene} proposed a benchmark using a thermal camera enabling easier hand detection without exploring its use for action recognition. In these previous works, hands are modeled using low-level features or intermediate representations following empirical grasp taxonomies \cite{bullock2015yale} and thus are limited compared to the 3D hand pose sequences used in this work. In \cite{rogez2015understanding}, synthetic hand poses are used to recognize grasps in static frames, whereas our interest is in dynamic actions and hand poses in real videos. From a hand pose estimation perspective, \cite{rogez20143d} proposed a small synthetic dataset of static poses and thus could not succesfully train data-hungry algorithms, recently relieved by larger synthetic datasets \citep{mueller2017real,choi2017robust}. Given that we also provide 6D object poses and 3D mesh models for a subset of objects, our dataset can be of interest to both object pose and joint hand-object tracking emerging communities \cite{tzionas2016capturing,sridhar2016real}. 
We compare our dataset with other first-person view datasets in Section \ref{dataset_comp}.

\textbf{RGB-D and pose-based action recognition:} Using depth sensors differs from traditional color action recognition in the fact that most successful color approaches \cite{wang2011action,CVPR2016_twostream} cannot be directly applied to the depth stream due to its nature: noisy, textureless and discontinuous pixel regions led to the necessity of depth-tailored methods. These methods usually focus on how to extract discriminative features from the depth images using local geometric descriptors \cite{hog2,hon4d, SNV} sensitive to viewpoint changes and view-invariant approaches \cite{hopc,novelview}. However, the recent trend is to take advantage of the depth channel to obtain robust body pose estimates \cite{shotton2013real} and use them directly as a feature to recognize actions, what is known as pose or skeleton action recognition. Popular approaches include the use of temporal state-space models \cite{xia2012view,wu2014leveraging,zhangefficient,wang2016graph,guicvpr17}, key-poses \cite{Zanfir_2013_ICCV,wang2016mining}, hand-crafted pose features \cite{vemulapalli2014human,rolling}, and temporal recurrent models \cite{du2015hierarchical,Veeriah_2015_ICCV,zhu2016co}. Having multiple data streams has led to the study of combining different sources of information such as depth and pose \cite{action3d,hog2,shahroudy2016multimodal,baek2016kinematic}, color and pose \cite{zhu2013fusing}, and all of them \cite{CVPR15_heterogeneous}. Most previous works in RGB-D action recognition focus on actions performed by the whole human body with some exceptions that are mainly application-oriented, such as hand gestures for human-computer interaction \cite{liu2013learning, hog2, nvidia,de2016skeleton,cviu16} and sign language \cite{wang2012robust}. Related to us, \cite{moghimi2014experiments} mounted a depth sensor to recognize egocentric activities and modeling hands using low-level skin features. Similar to our interests but in third-person view, \cite{yang2014cognitive,lei2012fine} used a hand tracker to obtain noisy estimates of hand pose in kitchen manipulation actions, while \cite{de2016skeleton} recognized basic hand gestures for human-computer interaction without objects involved. In these works, actions performed and pose labels are very limited due to the low quality of the hand tracker, while in this work we provide accurate hand pose labels to study more realistic hand actions. We go in depth and evaluate several baselines and state-of-the-art approaches in Sections \ref{evaluated} and \ref{experiments}.

\textbf{3D hand pose estimation:} Mainly due to the recent availability of RGB-D sensors, the field has made significant progress in object-less third-person view~\cite{oikonomidis2011efficient,keskin2012hand,tang2013real, Ionescu2014iter, liang2014parsing,qian2014realtime, neverova2015hand, oberweger2015training, sharp2015accurate, ye2016spatial} and more modest advances in first-person view~\cite{rogez20143d,oberweger2016efficiently,mueller2017real,choi2017robust}. In~\cite{oikonomidis2011full}, 3D tracking of a hand interacting with an object in third-person view was investigated.~\citep{hamer2010object} studied the use of object-grasp as hand pose prior, while~\citep{romero2013non} used the object shape as cue. An important limitation is the difficulty of obtaining accurate 3D hand pose annotations leading researchers to resort to synthetic~\cite{rogez20143d,sharp2015accurate,mueller2017real,choi2017robust,seungcvpr18}, manually or semi-automatically annotated~\cite{tang2014latent,tompson2014real, sun2015cascade,oberweger2016efficiently} datasets, resulting in non-realistic images, a low number of samples, and often inconsistent annotations. With the help of magnetic sensors for annotation and similar to~\citep{wetzler2015rule}, \cite{yuan2017bighand} proposed a big benchmark that included egocentric poses with no objects involved and showed that a ConvNet baseline can achieve state-of-the-art performance when enough training data is available. This was confirmed in a public challenge~\cite{yuan20172017}, also using a subset of our proposed dataset, and followed by a work~\cite{yuan20173d} analyzing the current state-of-the-art of the field.

\section{Daily hand-object actions dataset}

\subsection{Dataset overview}
The dataset contains 1,175 action videos belonging to 45 different action categories, in 3 different scenarios, and performed by 6 actors. A total of 105,459 RGB-D frames are annotated with accurate hand pose and action category. Action sequences present high inter-subject and intra-subject variability of style, speed, scale, and viewpoint. The object's 6-dimensional pose, 3D location and angle, and mesh model are also provided for 4 objects involving 10 different action categories. Our plan is to keep growing the dataset with more models and objects. In Fig. \ref{fig:dataset} we show some example frames for different action categories and hand-pose annotation visualization. 

\subsection{Hand-object actions}
\begin{figure*}[!t]
\begin{center}
    \includegraphics[width=\textwidth,trim={0 0mm 0 5mm}]{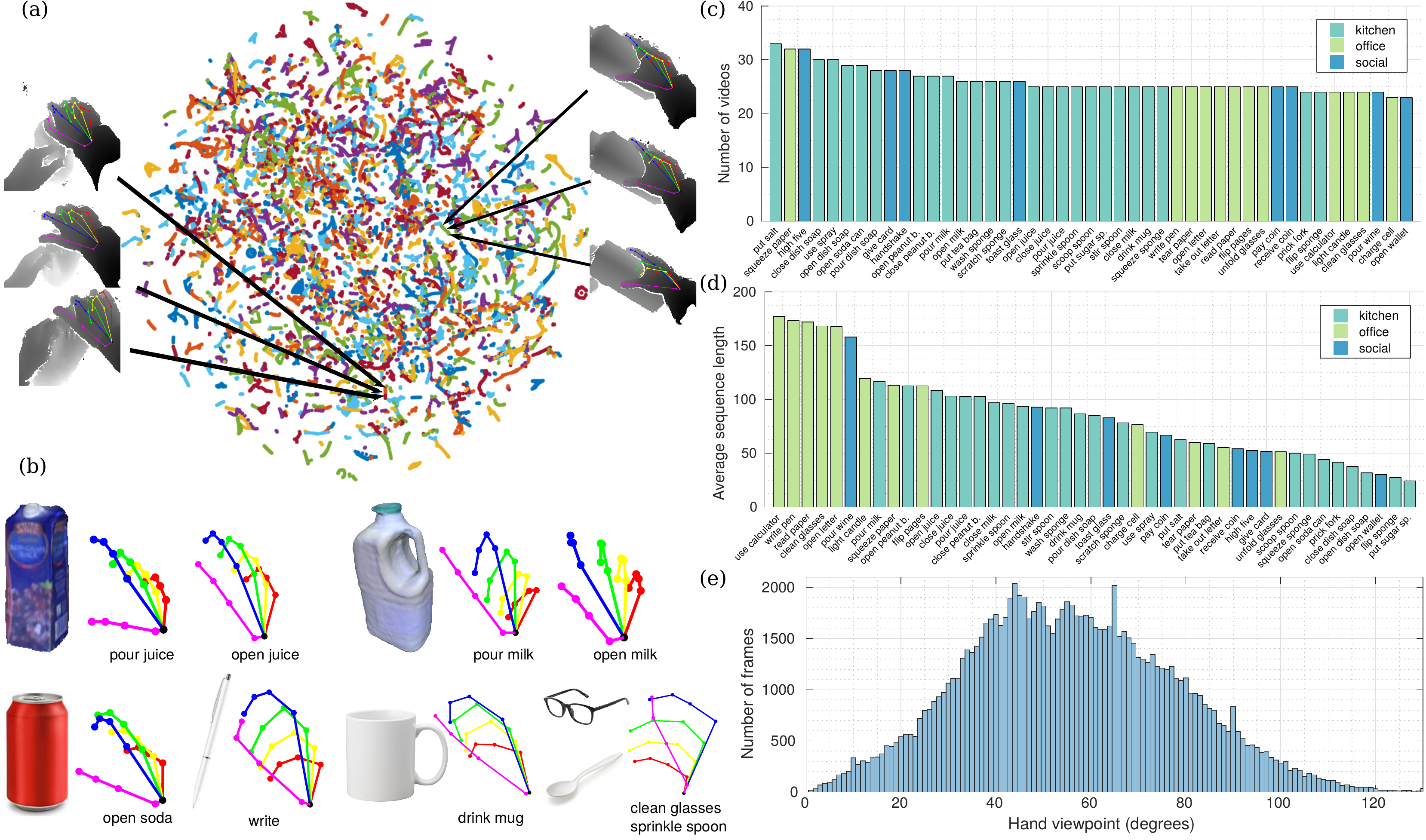}
   \caption{\textbf{(a)} t-SNE ~\cite{maaten2008visualizing} visualization of hand pose embedding over our dataset. Each colored dot represents a full hand pose and each trajectory an action sequence.
   \textbf{(b)} Correlation between objects, grasps, and actions. Shown poses are the average pose over all action sequences of a certain class. One object can have multiple grasps associated depending on the action performed (\eg, `juice carton' and `milk bottle') and one grasp can have multiple actions associated (\eg, lateral grasp present at `sprinkle' and `clean glasses'). \textbf{(c)} Number of action instances per hand action class. \textbf{(c)} Average number of frames in each video per hand action class. Our dataset contains both atomic and more temporally complex action classes. \textbf{(d)} Distribution of hand viewpoints, defined as angles between the direction of the camera and the direction of the palm of the hand.}
\label{fig:statistics}	
\end{center}
\end{figure*}

We captured 45 different daily hand action categories involving 26 different objects. We designed our action categories to span a high number of different hand configurations following the same taxonomy as~\cite{rogez2015understanding} and to be diverse in both hand pose and action space (see Fig. \ref{fig:statistics}). Each object has a minimum of one associated action (\eg, pen-`write') and a maximum of four (\eg, sponge-`wash', `scratch', `squeeze', and  flip'). These 45 hand actions were recorded and grouped in three different scenarios: kitchen (25), office (12) and social (8).  
In this work we consider each hand-object manipulation as a different action category similar to previous datasets~\cite{gtea}, although other definitions are possible~\cite{yang2014cognitive,wray2016sembed}.

\subsection{Sensors and data acquisition}
\textbf{Visual data:} We mounted an Intel RealSense SR300 RGB-D camera on the shoulder of the subject and captured sequences at 30 fps and resolutions 1920$\times$1080 and 640$\times$480 for the color and depth stream respectively.

\textbf{Pose annotation:} To obtain quality annotations of hand and object pose, the hand pose is captured using six magnetic sensors~\cite{trakSTAR} attached to the user's hand, five fingertips and one wrist, following~\cite{yuan2017bighand}. Each sensor provides position and orientation with 6 degrees of freedom and the full hand pose is inferred using inverse kinematics over a defined 21-joint hand model. Each sensor is 2 mm wide and when attached to the human hand does not influence the depth image. The color image is affected as the sensors and the tape attaching them are visible, however the hand is fully visible and actions distinguishable by using the color image. 
Regarding object pose, we attach one more sensor to the closest point to the center of mass that can be reached. 

\textbf{Recording process:} We asked 6 people, all right-handed, to perform the actions. 
Instructions on how to perform the action in a safe manner 
were given, however no instructions about style or speed were provided, in order to capture realistic data. Actions were labeled manually.

\begin{figure}[t]
\begin{center}
    \includegraphics[width=1\columnwidth]{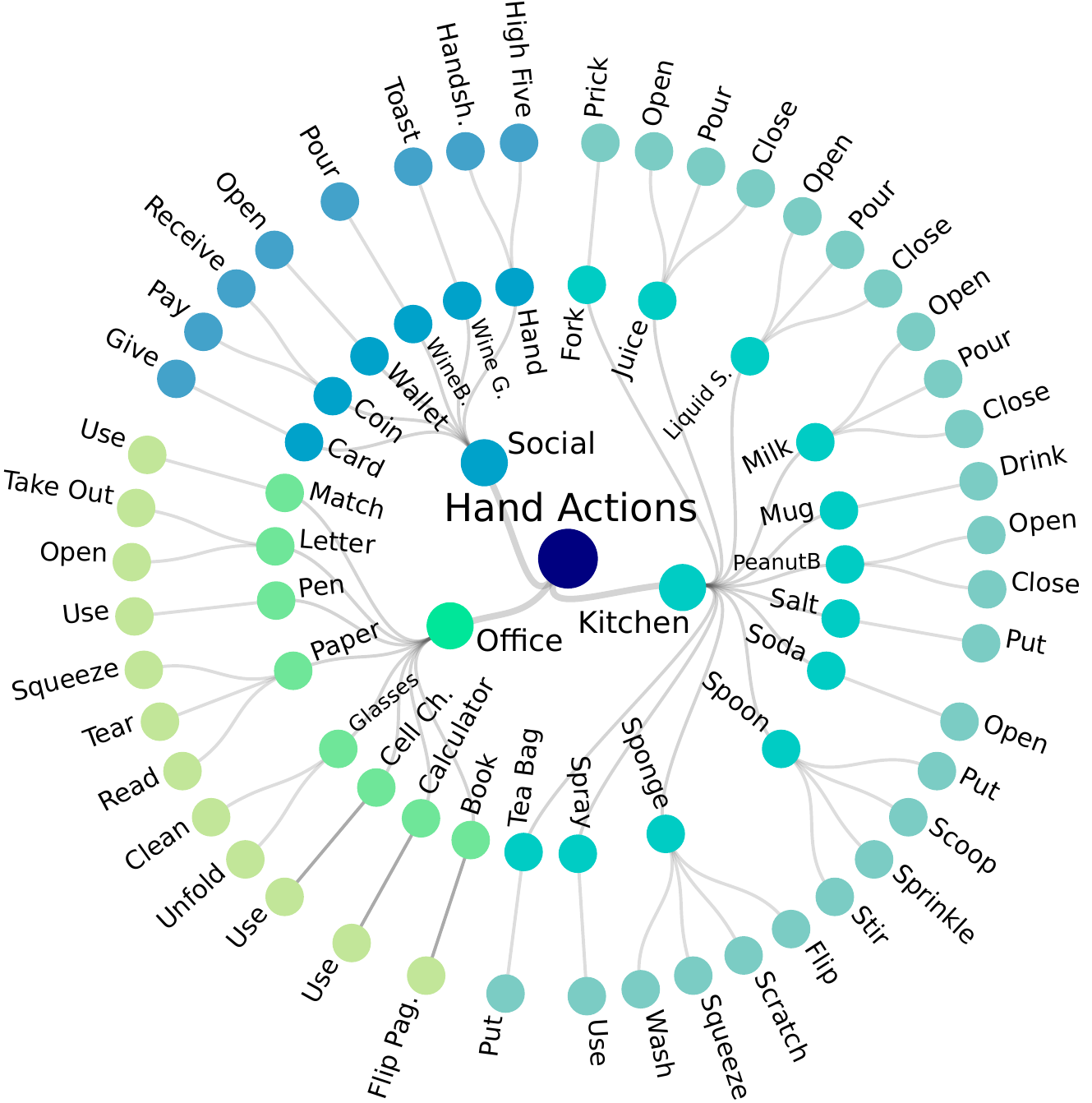}
    \caption{Taxonomy of our hand actions involving objects dataset. Some objects are associated with multiple actions (\eg, spoon, sponge, liquid soap), while some others have only one linked action (\eg, calculator, pen, cell charger).} 
	\label{pic:tree}
\end{center}
\end{figure}

\subsection{Dataset statistics}

\textbf{Taxonomy: }Fig. \ref{pic:tree}  shows the distribution of different actions per involved object. Some objects such as `spoon' have multiple actions (\eg, `stir',  `sprinkle', `scoop', and `put sugar'), while some objects have only one action (`use calculator'). Although it is not an object per se, we included `hand' as an object in actions `handshake' and `high five'.

\textbf{Videos per action class:} On average there are 26.11 sequences per class action and 45.19 sequences per object. For detailed per class numbers see Fig. \ref{fig:statistics} (c).

\textbf{Duration of videos:} Fig. \ref{fig:statistics} (d) shows the average number of video duration for the 45 action classes. Some action classes such as `put sugar' and `open wallet' involve short atomic movements, on average one second, while others such as `open letter' require more time to be executed. 

\textbf{Grasps:} We identified 34 different grasps following the same taxonomy as in~\cite{rogez2015understanding}, including the most frequently studied ones~\citep{caiunderstanding} (\textit{i.e.}, precision/power grasps for different object attributes such as prismatic/round/flat/deformable). In Fig. \ref{fig:statistics} (b) we show some examples of correlation between objects, hand poses, and actions.  

\textbf{Viewpoints}: In Fig. \ref{fig:statistics} (e) we show the distribution of frames per hand viewpoint. We define the viewpoint as the angle between the camera direction and the palm of the hand. The dataset presents viewpoints that are more prone to self-occlusion than typical ones in third-person view.

\textbf{Hand occlusion:} Fig. \ref{fig:visible_conf_pose} (a) (bottom) shows the average number of visible (not occluded by object or viewpoint) hand joints per action class. Most actions present a high degree of occlusion, on average 10 visible joints out of 21. 

\textbf{Object pose:} 6D object pose and mesh models are provided for the following objects involving 10 different actions: `milk bottle', `salt', `juice carton', and `liquid soap'.

\subsection{Comparison with other datasets} \label{dataset_comp}
In Table \ref{datasets} we summarize popular egocentric datasets that involve hands and objects in both dynamic and static fashion depending on their problem of interest. For conciseness, we have excluded from the table related datasets that do not partially or fully contain objects manipulations, \eg, \cite{pirsiavash2012detecting,oberweger2016efficiently,yuan2017bighand}. Note that previous datasets in action recognition~\cite{bambach2015lending,gtea,luoscene} do not include hand pose labels. On the other hand, pose and grasp datasets~\cite{rogez20143d,rogez2015understanding,bullock2015yale,cai2015scalable,choi2017robust,mueller2017real} do not contain dynamic actions and hand pose annotation is obtained by generating synthetic images or rough manual annotations~\citep{mueller2017real}. Our dataset `fills the gap' of egocentric dynamic hand action using pose and compares favorably in terms of diversity, number of frames, and use of real data.

\begin{table}[t]
  \centering
  \resizebox{\hsize}{!}{%
  \begin{tabular}{llccccc}
      \toprule
  Dataset &  Sensor & Real? &  Class. & Seq. & Frames & Labels \\
  \midrule
  Yale~\cite{bullock2015yale} & RGB & \checkmark & 33 & - & 9,100 & Grasp \\
  UTG~\cite{cai2015scalable} & RGB & \checkmark & 17 & - & - & Grasp \\
  GTEA~\cite{gtea} &  RGB & \checkmark & 61 & 525 & 31,222 & Action \\
  EgoHands~\cite{bambach2015lending} & RGB & \checkmark & 4 & 48 & 4,800 & Action \\
  \midrule
  GUN-71~\cite{rogez2015understanding} & RGB-D & \checkmark & 71 &- & 12,000 & Grasp \\ 
  UCI-EGO~\cite{rogez20143d} &  RGB-D & \xmark & - & - & 400 & Pose \\
  Choi \etal~\cite{choi2017robust} & RGB-D & \xmark & 33 &- & 16,500 & Grasp+Pose\\ 
  SynthHands~\cite{mueller2017real} & RGB-D & \xmark & - &- & 63,530 & Pose\\ 
  EgoDexter~\cite{mueller2017real} & RGB-D & \checkmark & - &- & 3,190 & Fingertips\\ 
  Luo \etal~\cite{luoscene} & RGB-D-T & \checkmark &  44 & 250 & 450,000 & Action \\

  \midrule
  Ours  & RGB-D & \checkmark  & 45 & 1,175 & 105,459 & Action+Pose\\
\bottomrule  

  \end{tabular}
    }
        \caption{First-person view datasets with hands and objects involved. Our proposed dataset is the first providing both hand pose and action annotations in real data (\textit{cf.} synthetic).}
          \label{datasets}
\end{table}

\section{Evaluated algorithms and baselines} \label{evaluated}
\subsection{Action recognition}

In order to evaluate the current state-of-the-art in action recognition we chose a variety of approaches that, we believe, cover the most representative trends in the literature as shown in Table \ref{table:50resultsGT}. As the nature of our data is RGB-D and we have hand pose, we focus our attention to RGB-D and pose-based action recognition approaches, although we also evaluate two RGB action recognition methods~\cite{CVPR2016_twostream,CVPR15_heterogeneous}. Note that, as discussed above, most of previous works in RGB-D action recognition involve full body poses instead of hands and some of them  might not be tailored for hand actions. We elaborate further on this in Section \ref{expaction}.

We start with one baseline to assess how the current state-of-the-art in RGB action recognition performs in our dataset. For this, and given that most successful RGB action recognition approaches~\cite{ma2016going,singh2016first} use ConvNets to learn descriptors from color and motion flow, we evaluate a recent two-stream architecture fine-tuned on our dataset~\cite{CVPR2016_twostream}.

About the depth modality, we first evaluate two local depth descriptor approaches, HOG$^2$~\cite{hog2} and HON4D~\cite{hon4d}, that exploit gradient and surface normal information as a feature for action recognition. As a global-scene depth descriptor, we evaluate the recent approach by~\cite{novelview} that learns view invariant features using ConvNets from several synthesized depth views of human body pose.

We follow our evaluation with pose-based action recognition methods. As our main baseline, we implemented a recurrent neural network with long-short term memory (LSTM) modules inspired in the architecture by~\cite{zhu2016co}. We also evaluate several state-of-the-art pose action recognition approaches. We start with descriptor-based methods such as Moving Pose~\cite{Zanfir_2013_ICCV} that encodes atomic motion information and~\cite{vemulapalli2014human}, which represents poses as points on a Lie group. For methods focusing on learning temporal dependencies, we evaluate HBRNN~\cite{du2015hierarchical}, Gram Matrix~\cite{zhangefficient} and TF~\cite{guicvpr17}. HBRNN consists of a bidirectional recurrent neural network with hierarchical layers designed to learn features from the body pose. Gram Matrix is currently the best performing method for body pose and uses Gram matrices to learn the dynamics of actions. TF learns both discriminative static poses and transitions between poses using decision forests.

To conclude, we evaluate one hybrid approach that jointly learns heterogeneous features (JOULE)~\cite{CVPR15_heterogeneous} using an iterative algorithm to learn features jointly taking into account all the data channels: color, depth, and hand pose. 

\subsection{Hand pose estimation}

To assess the state-of-the-art in hand pose estimation, we use the same ConvNet as~\cite{yuan2017bighand}. We choose this approach as it is easy to interpret and it was shown to provide good performance in a cross-benchmark evaluation~\cite{yuan2017bighand}. The chosen method is a discriminative approach operating on a frame-by-frame basis, which does not need any initialization and manual recovery when tracking fails~\cite{oikonomidis2011efficient,intel}.

\section{Benchmark evaluation results}\label{experiments}

\subsection{Action recognition} \label{expaction}

In the following we present our experiments in action recognition. In this section we assume the hand pose is given, \textit{i.e.}, we use the hand pose annotations obtained using the magnetic sensors and inverse-kinematics. We evaluate the use of estimated hand poses without the aid of the sensors for action recognition in Section \ref{handexp}. 

Following common practice in full body-pose action recognition~\cite{Zanfir_2013_ICCV,vemulapalli2014human}, we compensate for anthropomorphic and viewpoint differences by normalizing poses to have the same distance between pairs of joints and defining the wrist as the center of coordinates. 

\subsubsection{A baseline: LSTM}

We start our experimental evaluation with a simple yet powerful baseline: a recurrent neural network with long-short term memory module (LSTM). The architecture of our network is inspired by~\cite{zhu2016co} with two differences: we do not `go deep', and use a more conventional unidirectional network instead of bidirectional. Following~\cite{zhu2016co}, we set the number of neurons to 100 and a probability of dropout of $0.2$. We use TensorFlow and Adam optimizer. 

\textbf{Training and testing protocols:} We experiment with two protocols. The first protocol consists of using different partitions of the data for training and the rest for testing and we tried three different training:testing ratios of 1:3, 1:1 and 3:1 at sequence level. The second protocol is a 6-fold `leave-one-person-out' cross-validation, \textit{i.e.}, each fold consists of 5 subjects for training and one for testing. Results are presented in Table \ref{table:data_split}. We observe that following a cross-person protocol yields the worst results taking into account that in each fold we have similar training/testing proportions to the $3:1$ setting. This can be explained by the difference in hand action styles between subjects. In the rest of the paper we perform our experiments using the 1:1 setting with 600 action sequences for training and 575 for testing. 

\begin{figure*}[t]
\begin{center}
    \includegraphics[width=1\textwidth,trim={0 5mm 0 5mm}]{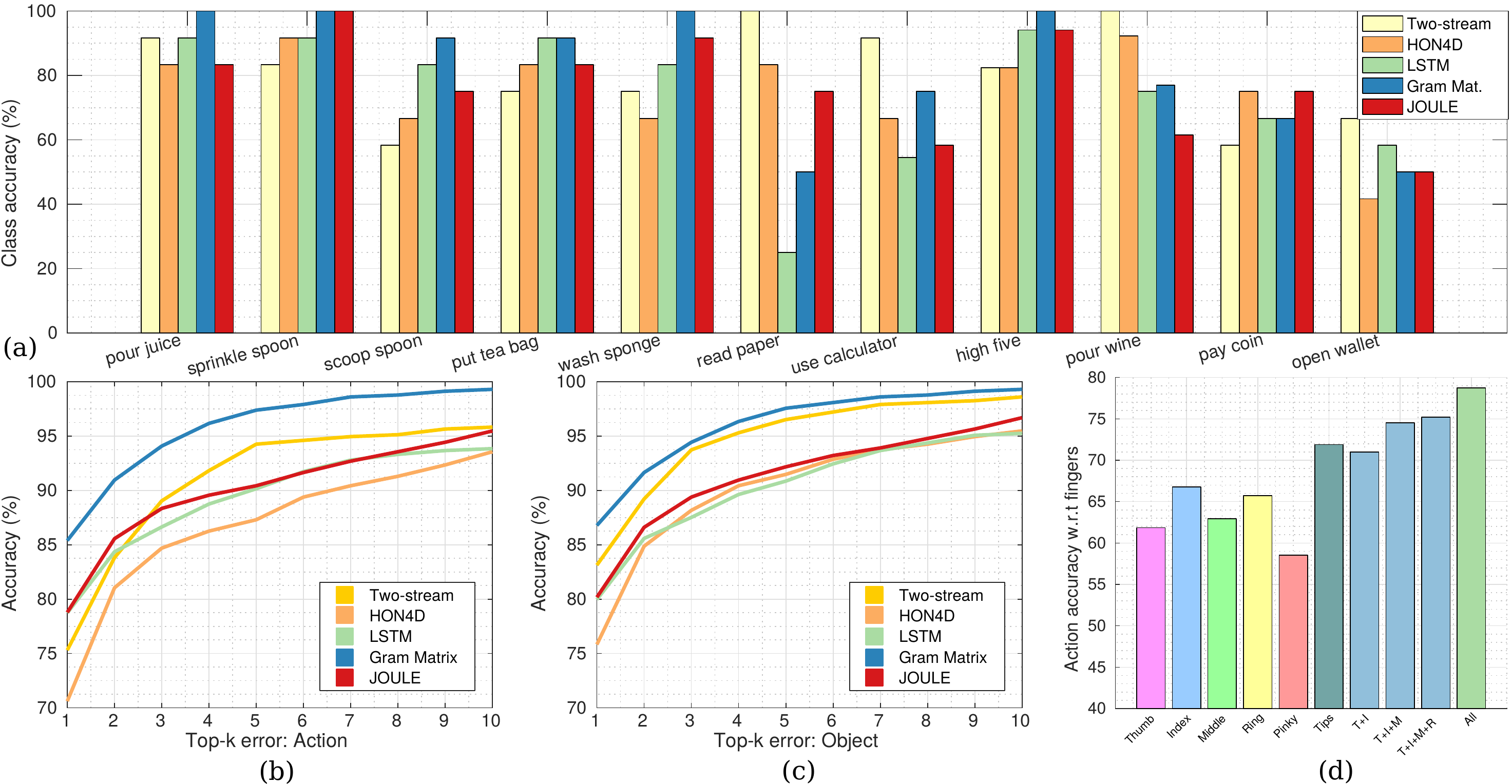}
    \caption{\textbf{(a)} Class accuracies of some representative methods on a subset of classes. \textbf{(b)} Top-$k$ action accuracy: true action label is in the top-$k$ action prediction hypothesis. \textbf{(c)} Top-$k$ object accuracy: manipulated object is in the top-$k$ action prediction hypothesis. \textbf{(d)} Impact of each of the five fingers, combinations of them, and fingertips on action recognition.} 
	\label{fig:compared_methods}
\end{center}
\end{figure*}

\textbf{Results discussion:} In Fig. \ref{fig:compared_methods} (a) we show the recognition accuracies per category on a subset actions and the action confusion matrix is shown in Fig. \ref{fig:visible_conf_pose} (b). Some actions such as `sprinkle spoon', `put tea bag' and `pour juice' are easily identifiable, while actions such as `open wallet' and `use calculator' are commonly confused, likely because hand poses are dissimilar and more subtle. In Fig. \ref{fig:compared_methods} (d) we show the contribution of each finger motion to action recognition performance, finding that the index is the most informative finger. Combining thumb and index poses boosts the accuracy, likely due to the fact that most grasps are explained by these two fingers~\citep{bullock2015yale}. Fingertips alone are also a high source of information due to being the highest articulated joints and being able to `explain' the hand pose.

\begin{table}
\begin{center}
\begin{tabular}{l|ccc|c}
\toprule
Protocol & 1:3 & 1:1 & 3:1 & cross-person \\
\midrule
Acc. (\%)  & 58.75 & 78.73 & 84.82 & 62.06 \\
\bottomrule
\end{tabular}
\end{center}
\caption{Action recognition results (percentage of correct video classification) for different training/testing protocols.} 
\label{table:data_split}
\end{table}

\subsubsection{State-of-the-art evaluation}

In Table \ref{table:50resultsGT} we show results for state-of-the-art approaches in different data modalities. We observe that the Two-stream~\cite{CVPR2016_twostream} method performs well when combining both spatial and temporal cues.  Depth methods tend to perform slightly worse than the rest of the methods, suggesting that they are not able to fully capture either the object cues or the hand pose. Note that for Novel View~\cite{novelview} we extracted deep features from a network trained on several synthetic views of bodies, which may not generalize well to hand poses  and fine-tuning in our dataset did not help. From all approaches, we observe that the ones using hand pose are the ones that achieve the best performance, with Gram Matrix~\cite{zhangefficient} and Lie group~\cite{vemulapalli2014human} performing particularly well, a result in line with the ones reported in body pose action recognition. 

In Fig. \ref{fig:compared_methods} we select some of the most representative methods and analyze their performance in detail. We observe that the pose method Gram Matrix outperforms the rest in most of the measures, specially when we retrieve the top $k$ action hypothesis (Fig. \ref{fig:compared_methods} (b)), showing the benefit of using hand pose for action recognition. Looking at Fig. \ref{fig:compared_methods} (a), we observe that Two-stream outperforms the rest of methods in some categories in which the object is big and the action does not involve much motion, \eg, `use calculator' and `read paper'. This good performance can be due to the pre-training of the spatial network on a big image recognition dataset. We further observe this in Fig. \ref{fig:compared_methods} (c) where we analyze the top $k$ hypothesis given by the prediction and look whether the predicted action contains the object being manipulated, suggesting that the network correctly recognizes the object but fails to capture the temporal dynamics.

\textbf{Hand pose vs. depth vs. color: } We performed one additional experiment using the JOULE~\cite{CVPR15_heterogeneous} approach by breaking down the contributions of each data modality. In Table \ref{table:50resultsGT} (bottom) we show that hand pose features are the most discriminative ones, although the performance can be increased by combining them with RGB and depth cues. This result suggests that hand poses capture complementary information to RGB and depth features.

\begin{table}
\begin{center}
\begin{tabular}{l|cc|c}
\toprule
Pose feature & Hand & Object & Hand+Object \\
\midrule
Action acc. (\%)  & 87.45 & 74.45 & 91.97  \\
\bottomrule
\end{tabular}
%}
\caption{We evaluate the use of 6D object pose for action recognition on a subset of our dataset. We observe the benefit of combining them with the hand pose. }
\label{table:object_pose}
\end{center}
\end{table}

\textbf{Object pose:} We did an additional experiment using the object pose as a feature for action recognition using the subset of actions that have annotated object poses: a total of 261 sequences for 10 different classes and 4 objects. We trained our LSTM baseline on half of the sequences and using three different inputs: hand pose, object pose, and both combined. In Table \ref{table:object_pose} we show the results and observe that  both object pose and hand pose features are complimentary and useful for recognizing egocentric hand-object actions.

\begin{table}[t]
  \centering
  \resizebox{\columnwidth}{!}{%
  \begin{tabular}{lccccc}
      \toprule
  Method & Year & Color & Depth & Pose & Acc. (\%) \\
  \midrule
    Two stream-color~\cite{CVPR2016_twostream} & 2016 & \checkmark & \xmark & \xmark & 61.56 \\
    Two stream-flow~\cite{CVPR2016_twostream} & 2016 & \checkmark & \xmark & \xmark & 69.91 \\
\midrule
    Two stream-all~\cite{CVPR2016_twostream} & 2016 & \checkmark & \xmark & \xmark & 75.30 \\

    \midrule
    HOG$^2$-depth~\cite{hog2} & 2013 & \xmark & \checkmark & \xmark &  59.83 \\
    HOG$^2$-depth+pose~\cite{hog2} & 2013 & \xmark & \checkmark & \checkmark &  66.78 \\

    HON4D~\cite{hon4d} & 2013 & \xmark & \checkmark & \xmark &  70.61 \\
	Novel View~\cite{novelview} & 2016 &\xmark & \checkmark & \xmark & 69.21 \\
    \midrule
    1-layer LSTM  & 2016 & \xmark & \xmark & \checkmark & 78.73 \\ 
    2-layer LSTM  & 2016 & \xmark & \xmark & \checkmark & 80.14 \\ 
    \midrule
    Moving Pose~\cite{Zanfir_2013_ICCV} & 2013 &  \xmark & \xmark & \checkmark & 56.34 \\
    Lie Group~\cite{vemulapalli2014human} & 2014 &  \xmark & \xmark & \checkmark & 82.69 \\
    HBRNN~\cite{du2015hierarchical} & 2015 & \xmark & \xmark & \checkmark & 77.40 \\ 
    Gram Matrix~\cite{zhangefficient} & 2016 & \xmark & \xmark & \checkmark & 85.39 \\ 
    TF~\cite{guicvpr17} & 2017 & \xmark & \xmark & \checkmark & 80.69 \\    
    \midrule
    JOULE-color~\cite{CVPR15_heterogeneous} & 2015 & \checkmark & \xmark & \xmark& 66.78 \\
    JOULE-depth~\cite{CVPR15_heterogeneous} & 2015 & \xmark & \checkmark & \xmark & 60.17 \\
    JOULE-pose~\cite{CVPR15_heterogeneous} & 2015 & \xmark & \xmark & \checkmark & 74.60 \\
\midrule
    JOULE-all~\cite{CVPR15_heterogeneous} & 2015 & \checkmark & \checkmark & \checkmark & 78.78 \\
\bottomrule

  \end{tabular}
    }
        \caption{Hand action recognition performance by different evaluated approaches on our proposed dataset.}
          \label{table:50resultsGT}
\end{table}

\subsection{Hand pose estimation}\label{handexp}
\begin{figure*}[t]
\begin{center}
    \includegraphics[width=\textwidth]{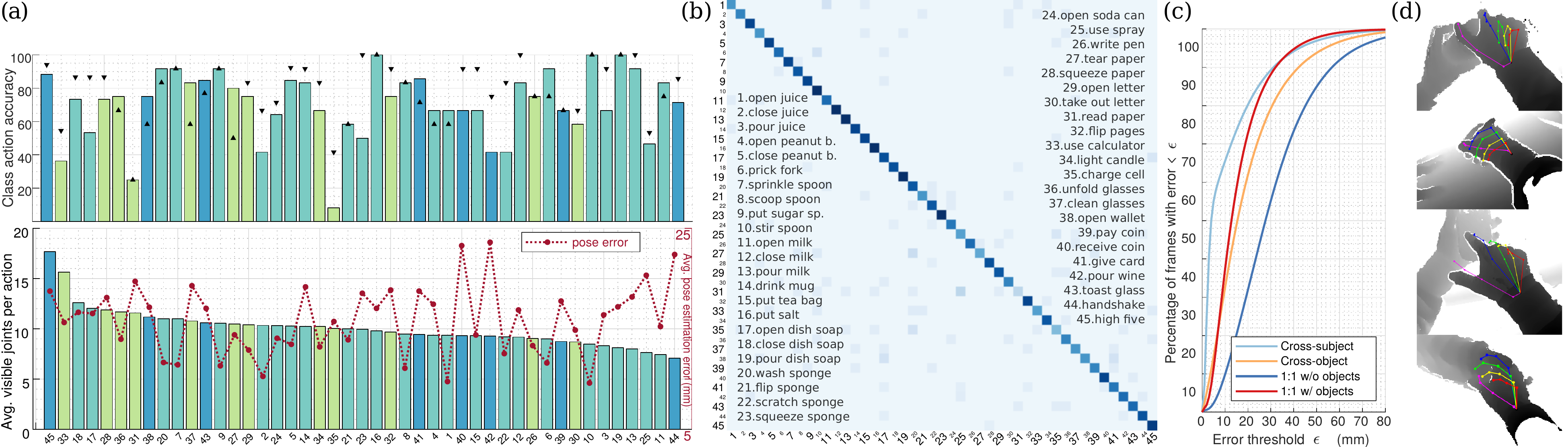}
    \caption{\textbf{(a)} \textit{Top: }Class action recognition accuracies for our LSTM baseline using estimated hand poses (accuracies with groundtruth poses are represented with black triangles). \textit{Bottom}: Average number of visible (not occluded) joints for hand actions on our dataset and its impact on hand pose estimation. \textbf{(b)} Hand action  confusion matrix for our LSTM baseline. \textbf{(c)}~Percentage of frames for different hand pose estimation error thresholds.  \textbf{(d)} Qualitative results on hand pose estimation.}
	\label{fig:visible_conf_pose}
\end{center}
\end{figure*}

\textbf{Training with objects vs. no objects:} One question raised while designing our experiments was whether we actually needed to annotate the hand pose in a close to ground truth accuracy to experiment with hand dynamic actions. We try to answer this question by estimating the hand poses of our hand action dataset in two ways partitioning our data as in our \textit{Action split}: using the nearly 300k \textit{object-free} egocentric samples from~\cite{yuan2017bighand} and using the images in the training set of our hand action dataset. As observed in Fig. \ref{fig:visible_conf_pose} (c) and Table \ref{table:action_pose}, the results suggest that having hand-object images in the training set is crucial to train state-of-the-art hand pose estimators, likely due to the fact that occlusions and object shapes need to be \textit{seen} by the estimator beforehand. To confirm this, we conducted two extra experiments: \textit{cross-subject} (half of the users in training and half in testing, all objects seen in both splits) and \textit{cross-object} (half of the objects in training and half in testing, all subjects seen in both splits). In Fig. \ref{fig:visible_conf_pose} (c) and Table \ref{table:action_pose} we observe that the network is able to generalize to unseen subjects but struggles to do so for unseen objects, suggesting that recognizing the shape of the object and its associated grasp is crucial to train hand pose estimators. This shows the need of having annotated hand poses interacting with objects and thus why our dataset can be of interest for the hand pose community.  In Fig. \ref{fig:visible_conf_pose} (d) we show some qualitative results in hand pose estimation in our proposed dataset and observe that, while not perfect, they are good enough for action recognition.

\textbf{Hand pose estimation and action recognition:} Now we try to answer the following key question: `how good is the current hand pose estimation for recognizing hand actions?'. In Table \ref{table:action_pose} we show results of hand action recognition by swapping the hand pose labels by the estimated ones in the test set. We observe that reducing the hand pose error by a factor of two yields a more than twofold improvement in action recognition. The difference in hand action recognition between using the hand pose labels and using the estimated ones in testing is 6.67\%. 
We also tested the two best performant methods from previous section, Lie group~\cite{vemulapalli2014human} and Gram Matrix~\cite{zhangefficient}. For Lie group we obtained an accuracy of 69.22\%, while for Gram Matrix a poor result of 32.22\% likely due to their strong assumptions in the noise distribution. 
On the other hand, our LSTM baseline shows more robust behavior in the presence of noisy hand pose estimates. In Fig. \ref{fig:visible_conf_pose} (a) we show how the hand occlusion affects the pose estimation quality and its impact on class recognition accuracies. Although some classes present a clear correlation between pose error and action accuracy degradation (\eg, `receive coin', `pour wine'), the LSTM is still able to obtain acceptable recognition rates likely due to being able to infer the action from temporal patterns and correctly estimated joints. For more insight, we analyzed the pose error per finger: T: 12.45, I: 15.48, M: 18.08, R: 16.69, P: 18.95, all in mm. Thumb and index joints present the lowest estimation error because of typically being less occluded in egocentric setting. According to previous section where we found that the motion from these two fingers was a high source of information, this can be a plausible explanation of why we can still obtain a good action recognition performance while having noisy hand pose estimates.  

\begin{table}
\centering
\resizebox{\columnwidth}{!}{%
\begin{tabular}{lcc}
\toprule
Hand pose protocol & Pose error (mm)  & Action (\%)  \\
\midrule
Cross-subject & 11.25 & - \\
Cross-object & 19.84 & - \\
\midrule
Action split (training w/o objects) & 31.03 & 29.63 \\
Action split (training w/ objects) & 14.34 & 72.06 \\
\midrule
Action split (GT mag.+IK poses)  & - & 78.73  \\
\bottomrule
\end{tabular}
}
\caption{Average hand pose estimation error, 3D distance over all 21 joints between magnetic poses and estimates, for different protocols and its impact on action recognition.}
\label{table:action_pose}
\end{table}

\section{Concluding remarks}
We have proposed a novel benchmark and presented experimental evaluations for RGB-D and pose-based, hand action recognition, in first-person setting. The benchmark provides both temporal action labels and full 3D hand pose labels, and additionally 6D object pose labels on a part of the dataset. Both RGB-D action recognition and 3D hand pose estimation are relatively new fields, and this is a first attempt to relate both of them similar to full human body. 
We have evaluated several baselines in our dataset and concluded that hand pose features are a rich source of information for recognizing manipulation actions. We believe that our dataset and experiments can encourage future work in multiple fields including action recognition, hand pose estimation, object pose estimation, and emerging ones, such as joint hand-object pose estimation. 

\textbf{Acknowledgements:} This work is part of Imperial College London-Samsung Research project, supported by Samsung Electronics.

{\small
\bibliographystyle{ieee}
\bibliography{main}
}

\end{document}